\pdfoutput=1

\documentclass[11pt]{article}

\usepackage[final]{acl}

\PassOptionsToPackage{hyphens}{url}

\usepackage[T1]{fontenc}
\usepackage[utf8]{inputenc}
\usepackage{times}
\usepackage{latexsym}
\usepackage{microtype}
\usepackage{inconsolata}

\usepackage{graphicx}
\usepackage{placeins}
\usepackage{tabularx}
\usepackage{natbib}
\usepackage{hyperref}
\usepackage{multirow}

\usepackage{array}   
\usepackage{ragged2e}     
\usepackage{booktabs}     
\usepackage{longtable}    
\usepackage{listings}     
\usepackage[hyphens]{url}
\usepackage{hhline}

\usepackage{graphicx}
\usepackage{float}        
\usepackage{tcolorbox}    
\usepackage{caption} 

\usepackage{float}
\usepackage{xcolor}

\usepackage{graphicx}

\usepackage{hyperref}  
\usepackage{makecell}

\usepackage{booktabs}
\usepackage{threeparttable} 

\usepackage{listings}
\usepackage{tcolorbox}
\usepackage{lscape}

\usepackage{longtable}
\usepackage{enumitem}

\title{Do LLMs Encode Frame Semantics? Evidence from Frame Identification}

\author{
  Jayanth Krishna Chundru\textsuperscript{1} \quad
  Rudrashis Poddar\textsuperscript{1} \quad
  Jie Cao\textsuperscript{2} \quad
  Tianyu Jiang\textsuperscript{1} \\
  \textsuperscript{1}University of Cincinnati    
  \textsuperscript{2}University of Oklahoma \\
  \texttt{\{chundrja, poddarrs\}@mail.uc.edu, \texttt{jie.cao@ou.edu}, tianyu.jiang@uc.edu} \\
}

\begin{document}
\maketitle
\begin{abstract}
We investigate whether large language models encode latent knowledge of frame semantics, focusing on frame identification, a core challenge in frame semantic parsing that involves selecting the appropriate semantic frame for a target word in context.~Using the FrameNet lexical resource,~we evaluate models under prompt-based inference and observe that they can perform frame identification effectively even without explicit supervision. To assess the impact of task-specific training, we fine-tune the model on FrameNet data, which substantially improves in-domain accuracy while generalizing well to out-of-domain benchmarks. Further analysis shows that the models can generate semantically coherent frame definitions, highlighting the model's internalized understanding of frame semantics.\end{abstract}

\section{Introduction}

Understanding the meaning of a word in context is a central challenge in natural language understanding, especially when words are polysemous and can evoke multiple meanings depending on usage. Frame semantics \citep{fillmore1976frame, fillmore1982frame} offers a structured approach to this problem by modeling meaning through frames, which represent typical situations or events along with the roles of the participants involved. The FrameNet 1.7 lexical resource \citep{ruppenhofer2016framenet} operationalizes this theory by associating over 13,000 lexical units with more than 1,200 semantic frames, each defining a distinct conceptual scenario with examples of how words trigger frames in context. Frame Semantic Parsing aims to automatically recover these frame-semantic structures from text, typically through three subtasks: target identification (detecting frame-evoking words or phrases), frame identification (determining the correct frame for each target), and argument identification (extracting and labeling the semantic roles, or frame elements). Our work focuses exclusively on frame identification involves selecting the appropriate semantic frame for a target word in context. For example, in the sentence:

\begin{quote}\textit{``In 1994, Pleasant Run \textbf{served} 346 children and 125 families.''}\end{quote}

The verb \textit{served} corresponds to multiple lexical units (LUs) in FrameNet, each representing a pairing of the word with a specific sense and an associated semantic frame. For example, \textit{serve.v} appears under the \textit{Capacity} frame---defined as ``have the capacity to serve a number of people (often said of meals or dishes)'', and also under the \textit{Assistance} frame---defined as “perform duties or services for someone.” In this context, the frame identification task requires to select the correct frame as \textit{Assistance}, as the verb refers to providing support services to children and families.

While traditional approaches to frame identification rely on supervised models and access to lexical disambiguation resources, we explore whether large language models (LLMs) 
inherently encode frame-semantic knowledge and can perform this task with minimal guidance. In this work, we evaluate the capabilities of LLMs to perform frame identification both through prompting and fine-tuning. We further analyze their semantic understanding through representational probing experiment.~Our code is open source and available online.\footnote{\url{https://github.com/cincynlp/FrameID}} In summary, our contributions include the following:
\begin{itemize}[itemsep=1pt,topsep=4pt]
    \item We demonstrate that prompting LLMs with simple and lightweight templates achieves strong performance in frame identification without any task-specific fine-tuning.
    \item We show that fine-tuning Llama-3.1-8B yields performance at par with state-of-the-art frame identification systems and generalizes well to two out-of-domain datasets.
    \item We probe the model’s latent frame knowledge by generating frame definitions and evaluating their effectiveness on Frame Identification.
\end{itemize}

\section{Related Work}

Frame identification - the task of determining the semantic frame evoked by a target word in context, has historically been approached with supervised learning on FrameNet annotations. Early systems such as \citealp{das-etal-2010-probabilistic} used feature-driven log-linear probabilistic models to perform frame identification, SEMAFOR \citep{chen-etal-2010-semafor} used log-linear models with rich syntactic and lexical features, while Open-SESAME \citep{swayamdipta2017frame} introduced a segmental RNN with a syntactic scaffold to jointly model frames and arguments. Subsequent work by \citet{hartmann-etal-2017-outofdomain} highlighted domain generalization issues, releasing the YAGS benchmark dataset, and \citet{peng-etal-2018-learning} proposed joint inference across disjoint datasets to further improve frame-semantic parsing.

Deep pre-trained models have catalyzed a shift in frame identification,  FIDO \citep{jiang-riloff-2021-exploiting} reframes the task as computing semantic similarity between a contextualized target embedding and definitions of candidate frames and lexical units. KGFI \citep{su-etal-2021-knowledge} enriches representations with FrameNet knowledge by incorporating definitions, frame elements, and frame–frame relations, projecting both targets and frames into a shared embedding space.~\citet{tamburini-2022-combining} combines the discriminatively pre-trained ELECTRA model with adaptive graph encoding of FrameNet information, yielding robust performance across benchmarks and evaluation settings. Recently, COFFTEA \citep{an-etal-2023-coarse} has employed a coarse to fine contrastive learning setup with dual encoders, improving target-frame alignment, retrieval efficiency and performance with or without lexical filtering. Although effective, these methods are primarily based on heavy task-specific supervision, exemplar sentences, and curated lexical mappings.

More recent studies have explored the capabilities of instruction-tuned large language models on structured semantic tasks such as semantic role labeling~\citep{cheng2024potential}, word sense disambiguation~\citep{basile2025exploring}, and AMR parsing~\citep{lee2023amr}. While some work has examined few-shot frame semantic parsing~\citep{shin-van-durme-2022-shot}, the ability of LLMs to perform FrameNet-style frame identification---particularly without any explicit task-specific fine-tuning remains underexplored. Prior approaches typically treat frame definitions as auxiliary input, rather than directly probing the latent frame-semantic knowledge that LLMs may encode. In contrast, our study examines whether LLMs can leverage their intrinsic semantic knowledge to identify frames in context.

\section{Methodology}
\label{sec:methodology}
We describe our approach to evaluate and improve frame-semantic understanding in LLMs, with a focus on the Frame Identification task.

\subsection{Inference-Time Prompting}
\label{ssec:infer-time-prompting}

We explore two prompt formats for Frame Identification using simple instructions, both designed to elicit direct, to-the-point answers from the model.

\paragraph{Simple Prompt:}~Presents the sentence, target word, and candidate frames (with definitions and lexical unit descriptions), asking the model to output the most appropriate \textit{frame name}.
\paragraph{Direct-QA Prompt:} Candidate frames are labeled (e.g., A, B, C), and the model \textit{selects the label} corresponding to the correct frame in a QA-style format.

Both prompt formats are evaluated under \textit{zero-shot} and \textit{few-shot} conditions (with 5 randomnly selected demonstration examples) are used to assess the model’s ability to leverage latent frame-semantic knowledge. These examples are selected from the training set to cover a variety of frames and target word usages, ensuring diversity in both lexical items and frame types. To enable automatic evaluation, we adopt structured output formats: \texttt{\{``frame\_name'': ``Causation''}\} for the Simple prompt and \texttt{\{``frame\_option'': ``A''}\} for the Direct QA prompt. We explored alternative prompting strategies (e.g., definition retrieval, rephrasing, chain-of-thought), but they did not yield significant gains. Thus, we focus on the Simple and Direct QA prompts, and conducted ablation studies~(\S\ref{ssec:ablation-study}).~See Table~\ref{tab:prompt-simple-directqa} for final prompt templates in appendix.

\subsection{QA Fine-Tuning}
\label{sec:qa-fine-tuning}
We fine-tune the model for contextual frame disambiguation by casting the task as question answering (QA). Each training instance consists of a sentence, a target word, and a list of candidate frames---each paired with its definition and lexical sense.~The candidates are labeled alphabetically (e.g., \texttt{A. Frame: Locale\_by\_use}, \texttt{B. Frame:Causation}, etc.), and the model is prompted to choose the correct label.

For fine-tuning, we compute logits over a restricted set of label tokens corresponding to frame choices using the model's language modeling head. The model is trained with cross-entropy loss to maximize the likelihood of the correct label at the next-token position. This setup encourages the model to resolve lexical ambiguity by selecting the frame that best aligns with the contextual meaning of the target word. The fine-tuning prompt is provided in Table~\ref{tab:prompt-qa-finetune} in the appendix.

\section{Experimental Results}
We evaluate prompting and fine-tuning for frame identification across in-domain and out-of-domain settings to assess their effectiveness with LLMs.
\label{sec:results}

\begin{table}[t]
\flushright
\begin{minipage}{0.49\textwidth}
\centering
\renewcommand{\arraystretch}{1.1}
\small
\begin{tabular}{llr}
\toprule
\textbf{Dataset} & \textbf{Model} & \textbf{Accuracy} \\
\midrule
\multirow{10}{*}{FN 1.5} 
  & \citeauthor{hartmann-etal-2017-outofdomain}~(\citeyear{hartmann-etal-2017-outofdomain}) & 87.6 \\
  & \citeauthor{yang-mitchell-2017-joint}~(\citeyear{yang-mitchell-2017-joint}) & 88.2 \\
  & Open-SESAME~(\citeyear{swayamdipta2017frame}) & 86.9 \\
  & \citeauthor{peng-etal-2018-learning}~(\citeyear{peng-etal-2018-learning}) & 90.0 \\
  & \citeauthor{jiang-riloff-2021-exploiting}~(\citeyear{jiang-riloff-2021-exploiting}) & 91.3 \\ 
   & KGFI (\citeyear{su-etal-2021-knowledge}) & 92.1 \\
  & \citeauthor{tamburini-2022-combining} (\citeyear{tamburini-2022-combining}) & \textbf{92.5} \\ 
  & COFFTEA (\citeyear{an-etal-2023-coarse}) & \textbf{92.5} \\ 
  & \citeauthor{devasier-etal-2024-robust} (\citeyear{devasier-etal-2024-robust}) & 91.7 \\ \cline{2-3}
  & Simple (zero-shot, Ours) & 82.4 \\
  & Simple (few-shot, Ours) & 82.7 \\
  & Direct-QA (zero-shot, Ours) & 82.5 \\
  & Direct-QA (few-shot, Ours) & \underline{83.3} \\
  & QA Fine-Tuning (Ours) & \underline{91.7} \\
\midrule
\multirow{8}{*}{FN 1.7} 
  & \citeauthor{peng-etal-2018-learning}~(\citeyear{peng-etal-2018-learning}) & 89.1 \\
  & \citeauthor{jiang-riloff-2021-exploiting}~(\citeyear{jiang-riloff-2021-exploiting}) & 92.1 \\ 
  & KGFI (\citeyear{su-etal-2021-knowledge}) & 92.4 \\
  & \citeauthor{tamburini-2022-combining} (\citeyear{tamburini-2022-combining}) & 92.3 \\ 
  & COFFTEA (\citeyear{an-etal-2023-coarse}) & \textbf{92.6} \\
   & \citeauthor{devasier-etal-2024-robust} (\citeyear{devasier-etal-2024-robust}) & 92.3 \\ \cline{2-3}
  & Simple (zero-shot, Ours) & 80.0 \\
  & Simple (few-shot, Ours) & 80.9 \\
  & Direct-QA (zero-shot, Ours) & 81.7 \\
  & Direct-QA (few-shot, Ours) & \underline{83.5} \\
  & QA Fine-Tuning (Ours) & \underline{91.9} \\
\bottomrule
\end{tabular}
\caption{Accuracy comparison for Frame Identification on FN 1.5 and FN 1.7 datasets (avg. over 3 runs).}
\label{tab:frame-identification}
\end{minipage}
\end{table}

\subsection{In-Domain Evaluation}
We evaluate on FrameNet~(FN) 1.5 and 1.7, which provide sentence-level annotations that link lexical units (LU)---context-sensitive word senses annotated with the frames they evoke~\citep{baker1998berkeley}. For example, the LU  \texttt{serve} may evoke the \textit{Assistance} frame when referring to helping others, or the \textit{Capacity} frame when referring to portion sizes. FN 1.7 expands FN 1.5 with 20\% more annotated examples, more defined frame definitions and increased lexical diversity. We adopt the standard data split used by \citet{das2014framesemantic} for FN 1.5 and by \citet{swayamdipta2017frame} for FN 1.7.

Table \ref{tab:frame-identification} compares our method with prior and state-of-the-art models, FIDO \citep{jiang-riloff-2021-exploiting} introduced definition matching by modeling similarity between contextualized targets and frame/lexical unit definitions.~\citet{su-etal-2021-knowledge} extended this with richer FrameNet definitions, frame elements, and frame relations in a shared embedding space.~\citet{tamburini-2022-combining} combined ELECTRA with adaptive graph encoding of FrameNet, training on full text annotations and examining the role of exemplar sentences, though these did not yield consistent improvements.~COFFTEA~\citep{an-etal-2023-coarse} leveraged exemplar data, in combination with a coarse-to-fine dual encoder trained with contrastive learning, resulting in improved frame–target alignment and modest overall performance gains.~\citet{devasier-etal-2024-robust} introduced lexical unit prefix trees and negative sampling to improve frame identification, especially on the rare frames.

In contrast, we evaluate Llama-3.1-8B-Instruct using lightweight prompting strategies under zero-shot and few-shot settings (\S\ref{ssec:infer-time-prompting}). Few-shot Direct QA achieves the strongest results, with overall accuracies of 83.3\% on FN~1.5 and 83.5\% on FN~1.7, demonstrating the model’s solid frame understanding without task-specific supervision. Furthermore, fine-tuning Llama-3.1-8B on FN 1.5 and FN 1.7 data without using exemplars yields accuracies of 91.7\% and 91.9\%, respectively, highlighting the model’s solid frame understanding, putting it on par with current state-of-the-art models.

For fine-tuning, we train the base Llama-3.1-8B (base model) using LoRA~\citep{hu2021lora} rank of 16, lora\_alpha set to 32, performed with a batch size of 1, over 3 epochs, using a learning rate of 2e-5 and mixed-precision (fp16).
\renewcommand{\arraystretch}{1.2}

\begin{table}[t]
\centering
\small
\setlength{\tabcolsep}{2pt}
\begin{tabular}{@{}lcc@{}}
\toprule
\textbf{Model} & \textbf{YAGS (\%)} & \textbf{Artifacts (\%)} \\
\midrule
\citet{hartmann-etal-2017-outofdomain} & 62.5 & -- \\
FIDO \citep{jiang-riloff-2021-exploiting} & 70.5 & -- \\
\midrule
Llama-3.1-8B (Zero-Shot) & 65.4 & 25.6 \\
Llama-3.1-8B (QA Fine-Tuning) & \textbf{80.7} & \textbf{49.6} \\
\bottomrule
\end{tabular}
\caption{Out-of-domain accuracy on YAGS and Artifacts(avg. over 3 runs).}
\label{tab:ood-results}
\end{table}

\subsection{Out-of-Domain Evaluation}
To assess generalization beyond the training distribution, we evaluated the FN~1.7 fine-tuned model~(\S\ref{sec:qa-fine-tuning}) on the two established out-of-domain datasets.~\textbf{YAGS}~\citep{hartmann-etal-2017-outofdomain} is a benchmark annotated with FN~1.5 frames, derived from user-generated posts on the Yahoo!~Answers online forum, including unknown targets~(not linked to any LU in FN 1.5) and unlinked targets~(gold frames not among the target's FrameNet associated frames), making it a strong test of robustness.~\textbf{Artifacts}~\citep{jiang-riloff-2021-learning} consists of 938 physical objects annotated with FrameNet frames representing their prototypical functions. Unlike sentence-level datasets such as FrameNet and YAGS, Artifacts provides entity mentions individually and asks the model to pick one frame that best represents how it is typically used. This introduces a structural shift from sentence-level to phrase-level reasoning, thereby probing whether models can abstract frame knowledge away from contextual information.

Table~\ref{tab:ood-results} shows that the FN~1.7 fine-tuned model achieves 80.7\% on YAGS benchmark outperforming both FIDO \citep{jiang-riloff-2021-exploiting} and the zero-shot Llama-3.1-8B baseline, and improves from 25.6\% to 49.6\% on Artifacts. These results highlight the model's ability to generalize across domains and input formats.

\renewcommand{\arraystretch}{1.2}
\begin{table}[t]
\centering
\small
\resizebox{\columnwidth}{!}{%
\begin{tabular}{lcc}
\toprule
\textbf{Prompt Type (Granularity)} & \textbf{Zero-Shot} & \textbf{Few-Shot} \\
\midrule
Simple (Frame Names)                & 59.9 & 79.1 \\
Simple (Frame Names \& Defs)                 & 76.2 & 79.8 \\
Simple (Frame Names \& LU Defs)     & 76.5 & 80.9 \\
Simple (Frame Names, Defs \& LU Defs)      & 80.0 & 80.9 \\
\midrule
Direct-QA (Frame Names)             & 80.1 & 81.3 \\
Direct-QA (Frame Names \& Defs)              & 80.6 & 80.8 \\
Direct-QA (Frame Names \& LU Defs)  & 80.8 & \textbf{83.5} \\
Direct-QA (Frame Names, Defs \& LU Defs)   & \textbf{81.7} & 82.8 \\
\bottomrule
\end{tabular}%
}
\caption{ Prompting strategy and input granularity ablation on FN 1.7 (avg. over 3 runs).}
\label{tab:analysis-table}
\end{table}
\subsection{Ablation Study}
\label{ssec:ablation-study}
We perform an ablation study on FN~1.7 by systematically varying both the prompt types (\textit{Simple} vs.\ \textit{Direct QA}) and the input granularity (with or without frame and lexical unit (LU) definitions). As shown in Table~\ref{tab:analysis-table}, comparing the bottom 2 rows with upper 2 rows in each prompt type, adding LU definitions consistently improves accuracy in both prompt types. Direct QA performs better than Simple prompt, with the best few-shot result (83.5\%) achieved using frame names and LU definitions. Interestingly, we also observe that frame names occasionally outperform full definitions, suggesting the model favors concise, unambiguous semantic cues over longer and descriptive definitions.
\begin{table}[t]
\centering
\footnotesize
\resizebox{0.9\linewidth}{!}{
\begin{tabular}{l r}
\toprule
\textbf{Error Category} & \textbf{Count} \\
\midrule
FIDO wrong predictions           & 519 \\
Llama-3.1-8B wrong predictions          & 538 \\
\midrule
Common wrong predictions         & 320 \\
\quad  Agreeing wrong predictions    & \textbf{296} \\
\quad  Disagreeing wrong predictions & 24 \\
\bottomrule
\end{tabular}
}
\caption{Error breakdown of FIDO and Llama-3.1-8B, including overlap and disagreement.}
\label{tab:error-overlap}
\end{table}

\begin{table}[t]
\centering
\resizebox{0.9\linewidth}{!}{
\begin{tabular}{@{}lcc@{}}
\toprule
\textbf{Model} & \textbf{Zero-Shot} & \textbf{Few-Shot} \\
\midrule
Llama-3.1-8B-Instruct    & 75.0 & 75.4 \\
Deepseek-V3.1   & 79.3 & 79.1 \\
GPT-4o   & 80.0 & 80.1 \\
\midrule
FrameNet 1.7 & 78.4 & 79.3 \\
\bottomrule
\end{tabular}}
\caption{Frame identification accuracy when replacing gold FN 1.7 definitions with LLM-generated definitions.}
\label{tab:Llama_generated_defintions}
\end{table}

\definecolor{darkgreen}{rgb}{0.0, 0.5, 0.0}
\begin{table*}[ht!]
    \centering
    \begin{tabularx}{\linewidth}{|p{0.30\linewidth}|X|}
        \hline
        \textbf{Sentence} & \textbf{Candidate Frames - Frame \& Lexical Unit definitions} \\
        \hline
        ``However, aetna's employee benefits division, which includes its \textcolor{blue}{group} health insurance operations, posted a 34\% profit gain to \$ 106 million.'' &
        \textcolor{darkgreen}{\textit{Organization}} - This frame describes intentionally formed human social groups (here termed Organizations) with .... - group: an organized set of individuals set upon some task. \vspace{0.3em}\par
        \textcolor{red}{\textit{Aggregate}} - This frame contains nouns denoting Aggregates of Individuals.  The Aggregates may be described by .... - group: a number of people or things located, gathered, or classed together.\\
        \hline
        ``a syria - eu trade accord hurdle was resolved in october with \textcolor{blue}{agreement} on a wmd clause, subject to final approval by eu foreign ministers.'' &
        
        \textcolor{darkgreen}{\textit{Be\_in\_agreement\_on\_action}} - Two (or more) people (the Parties, also encodable as Party\_1 and Party\_2) have an agreement.... - agreement: negotiated and typically legally binding arrangement.
        \vspace{0.3em}\par
        \textit{Be\_in\_agreement\_on\_assessment} -The Cognizers have a similarity (or dissimilarity) in their Opinion... - agreement: accordance in opinion or feeling.
        \vspace{0.3em}\par
        \textit{Documents} -  Words in the frame refer to any Document that has a legal status or conventional social significance... - agreement: a contract by which one party conveys land, property, services, etc. to another for a specified time.
        \vspace{0.3em}\par
        \textcolor{red}{\textit{Make\_agreement\_on\_action}}- Two (or more) people (the Parties, also encodable as Party\_1 and Party\_2) negotiate an agreement... - agreement.n: a negotiated and typically legally binding arrangement. \\
        \hline
        ``upon completion of several uranium exploration projects ,syria began experiments to \textcolor{blue}{extract} uranium from its vast phosphoric rock reserves .'' & 
        \textcolor{darkgreen}{\textit{Removing}} -  An Agent causes a Theme to move away from a location, the Source... - extract: remove, especially by effort or force.
        \vspace{0.3em}\par
        \textcolor{red}{\textit{Mining}} - A Miner attempts to obtain a desirable Resource, rocks and minerals, located in a... - extract: the process of removing resources from the earth.\\
        \hline
    \end{tabularx}
    \caption{Sample agreeing wrong predictions, where both FIDO and Llama-3.1-8B predict the same incorrect frame. Target words are marked in \textcolor{blue}{blue}, gold frames in \textcolor{darkgreen}{green}, and predicted frames in \textcolor{red}{red}.}
    \label{tab:sample-agreeing-wrong-predictions}
\end{table*}

\subsection{Error Comparison with FIDO}
To compare model behavior, we analyzed errors made by FIDO~\citep{jiang-riloff-2021-exploiting} and our Llama-3.1-8B fine-tuned on FN~1.7.~FIDO produced 519 error predictions, while~Llama-3.1-8B had 538 errors, with 320 overlapping. Of these, 296 were \textit{agreeing wrong predictions} (same incorrect frame) and 24 \textit{disagreeing wrong predictions} (different incorrect frames). Sample examples are shown in Table~\ref{tab:sample-agreeing-wrong-predictions}.~These overlaps reveal shared confusion, reflecting subtleties in the frame inventory and target lexical senses that challenge frame identification systems.

\subsection{LLM-Generated Definitions}

We extend our analysis by examining whether large language models encode frame-semantic knowledge in an inherent manner.~Specifically, we prompt LLMs to generate definitions for FrameNet~1.7 frames using only the frame names as input, thereby removing any reliance on additional lexical information. 

While related work such as \citet{han-etal-2024-definition} also generates frame definitions, their goal is instead to create definitions for \textit{induced} frames in order to make unsupervised clusters interpretable and usable as lexical resources.~In contrast, our use of definition generation is diagnostic:~we probe whether the internal knowledge of LLMs about FrameNet frames can be surfaced through definition generation.~To evaluate the~quality of these generated definitions, we assess their extrinsic utility on the Frame Identification task by replacing gold definitions with LLM-generated definitions in the Direct QA format (without revealing frame names), while fixing the inference model to Llama-3.1-8B-Instruct for consistency.~As reported in Table~\ref{tab:Llama_generated_defintions}, the resulting accuracy on the Frame Identification remains comparable to that achieved with gold definitions, showing that the generated definitions capture sufficient semantic content to support frame disambiguation.

\section{Conclusion}

\noindent
We examined whether large language models (LLMs) encode the frame semantics knowledge required for Frame Identification.~Prompting Llama-3.1-8B-Instruct achieves competitive performance relative to fine-tuned models, even in both zero-shot and few-shot settings.~Fine-tuning the Llama further improves results, reaching the prior state-of-the-art performance on FrameNet benchmarks.~Evaluation of the FN 1.7 fine-tuned model on the two out-of-distribution datasets~(\textit{YAGS} and \textit{Artifacts}) demonstrates the model’s ability to generalize across domains and input formats. Further analysis showed that LLMs can also generate coherent frame definitions, which produce comparable results on the Frame Identification task. Overall, these findings demonstrate that large language models encode frame-semantic knowledge and can serve as effective solutions for frame-semantic tasks with minimal supervision.

\section*{Limitations}

We acknowledge several limitations of our study. Our comprehensive experiments are confined to the Llama family of language models and our analysis is confined to English and FrameNet-style frame inventories. It  remains an open question to extend LLM-based frame identification to multilingual contexts~\citep{baker-etal-2018-frame}, alternative frame ontologies~\citep{pradhan-etal-2022-propbank}, or broader frame-driven language systems~\citep{bobrow1977gus,lassila2001role,cao-zhang-2021-comparative}. Our analysis is confined to the Frame Identification task, with the other key components of Frame Semantic Parsing left unaddressed in this study. Our quantitative evaluation of LLM-generated frame definitions is limited to their impact on the Frame Identification task; a more rigorous human annotation-based evaluation would provide deeper insights. Moreover, our current approach to definition generation employed only a zero-shot prompting style, which already achieved strong performance when compared against gold FrameNet definitions. Nonetheless, exploring diverse prompting strategies (e.g., structured scaffolds, exemplars, or chain-of-thought prompts) may further enhance the quality of the definition.

\section*{Acknowledgements}

We thank the CincyNLP group for their helpful comments and the anonymous EMNLP reviewers for their valuable feedback and suggestions.

\bibliography{custom}

\twocolumn
\appendix
\section{Dataset Statistics}

Table~\ref{tab:dataset-sizes} summarizes the number of examples in each dataset split used in our experiments.~For FN~1.5, we adopt the standard splits from~\citet{das2014framesemantic}, which include 15,017 training, 4,463 validation,  and 4,457 test examples. For FN~1.7, we use the splits from \citet{swayamdipta2017frame}, comprising 19,391 training, 2,272 validation, and 6,714 test examples.
\label{sec:datases}
\begin{table}[H]  
\centering
\begin{tabular}{lccc}
\toprule
\textbf{Dataset} & \textbf{Train} & \textbf{Dev} & \textbf{Test} \\
\midrule
FrameNet 1.5 & 15,017 &  4,463 &  4,457 \\
FrameNet 1.7 & 19,391 & 2,272 & 6,714 \\
YAGS         & --     & 944 & 1971 \\
Artifacts    & --     & --    & 938   \\
\bottomrule
\end{tabular}
\caption{Number of examples in each dataset split.}
\label{tab:dataset-sizes}
\end{table}
To evaluate out-of-domain generalization, we use two datasets: YAGS and Artifacts. The YAGS dataset \citep{hartmann-etal-2017-outofdomain}, derived from user-generated content on Yahoo! Question Answers posts, including 944 validation, and 1971 test examples. It introduces domain shift and challenging lexical ambiguity. The Artifacts dataset \citep{jiang-riloff-2021-learning} consists of 938 noun phrases annotated with FrameNet frames, targeting the prototypical functions of physical objects. It differs structurally from sentence-level FrameNet inputs and is used solely for zero-shot evaluation.

\section{Generalization and Additional Analyses}
We conducted two additional analyzes to assess the robustness and generality of our findings.
\paragraph{Ambiguous Cases.}
Ambiguous cases are instances where a target word can plausibly evoke multiple frames in FrameNet, making disambiguation particularly challenging (e.g., \textit{serve} evoking either the \textit{Capacity} or \textit{Assistance} frame). As shown in Table~\ref{tab:direct-prompts}, incorporating Lexical Unit (LU) definitions provides a clear advantage: the best performance is achieved with \textit{Frame Names \& LU Defs} in the few-shot setting and both the variants using LU defs in the zero-shot setting.~Overall, LU definitions act as strong semantic cues that stabilize performance in case of frame ambiguity.

\begin{table}[H]
\begin{flushright}
\resizebox{\columnwidth}{!}{%
\begin{tabular}{llccc}
\toprule
\multirow{2}{*}{\textbf{Prompt Type (Granularity)}} & 
\multicolumn{2}{c}{\textbf{Ministral}} & 
\multicolumn{2}{c}{\textbf{Qwen}} \\
\cmidrule(lr){2-3} \cmidrule(lr){4-5}
 & \textbf{0-Shot} & \textbf{Few-Shot} & \textbf{0-Shot} & \textbf{Few-Shot} \\
\midrule
 Simple (Frame Names)               & 77.4 & 80.0 & 83.0 & 83.1 \\
 Simple (Frame Names \& Defs)               & 78.0 & 80.1 & 81.9 & 82.0 \\
 Simple (Frame Names \& LU Defs)   & \textbf{82.1} & \textbf{81.9} & \textbf{83.4} & 83.5\\
 Simple (Frame Names, Defs \& LU Defs)    & 79.2 & \textbf{81.9} & 81.6 & 82.4 \\
\midrule
 Direct-QA (Frame Names)             & 78.7 & 76.2 & 82.0 & 82.6 \\
 Direct-QA (Frame Names \& Defs)               & 78.0 & 79.2 & 82.2 & 81.0 \\
 Direct-QA (Frame Names \& LU Defs)    & 80.5 & 80.8 & 82.9 & \textbf{83.7} \\
 Direct-QA (Frame Names, Defs \& LU Defs )    & 81.1 & 81.1 & 83.0 & 83.3 \\
\bottomrule
\end{tabular}%
}
\caption{Prompting strategy and input granularity on FN~1.7 (avg. over 3 runs).}
\label{tab:analysis-table-othermodels}
\end{flushright}
\end{table}

\begin{table}[t]
\begin{flushright}
\resizebox{\columnwidth}{!}{%
\begin{tabular}{lcc}
\toprule
\textbf{Granularity} & \textbf{Zero-shot} & \textbf{Few-shot} \\
\midrule
Frame Names             & 65.1 & 66.6 \\
Frame Names \& Defs              & 64.6 & 67.6 \\
Frame Names \& LU Defs    & 65.6 & \textbf{69.7} \\
Frame Names, Defs \& LU Defs     & 67.0 & 65.0 \\
\bottomrule
\end{tabular}%
}
\caption{Performance of Direct-QA prompt on ambiguous cases on Llama-3.1-8B-Instruct.}
\label{tab:direct-prompts}
\end{flushright}
\end{table}

\paragraph{Generalization Across LLMs.}
To verify that our prompting strategies extend beyond Llama, we evaluate two open-source instruction-tuned models:~Ministral-8B-Instruct-2410 and Qwen-3-8B.~Results from Table~\ref{tab:analysis-table-othermodels} shows that our lightweight prompting strategies generalize well across models.~Qwen consistently achieves the stronger results, while Ministral demonstrates competitive performance. Few-shot prompting yields only marginal gains, over zero-shot prompting.~Across model families, both Simple and Direct-QA prompts remain effective, with combinations of frame names and LU definitions yielding the highest accuracy. These findings confirm that structured prompts are broadly effective for frame identification, enabling LLMs to perform competitively without extensive exemplars or fine-tuning.

\section{Qualitative Definition Analysis}
\label{sec:qualitative-def}
\begin{table*}[t]
\centering
\begin{tabularx}{\textwidth}{|p{0.15\textwidth}|X|X|}
\hline
\textbf{Frame Name} & \textbf{Gold Definition} & \textbf{Llama Generated Definition} \\
\hline
Activity\_prepare & `` An Agent prepares for an Activity. The troops were preparing themselves for the mission. The boats are getting ready to leave the pier. '' & `` The process of making something or someone ready for a specific activity or situation, often involving physical or mental preparation, planning, or setup. '' \\
\hline
Judicial\_body & `` A  Source\_of\_authority, typically a government, relies upon a set of institutions specialized for the administration of justice. These Judicial\_body s hear cases  that originate within their Jurisdiction and relate  to the Task given to them by the Source\_of\_authority. Often, a Judicial\_body is part of a larger system  within which its Rank can be specified. The Judicial\_body is typically located in a specific location, the Seat.   'The federal judiciary's certification process is the foundation ..... '' & `` A government institution or agency responsible for administering justice, such as a court, tribunal, or council, that has the authority to hear and decide cases, interpret laws, and enforce the law. '' \\
\hline
Measure\_mass & `` This frame is concerned with nouns that denote Units of measurement for masses. The Frame Elements are Unit, Count,  and Stuff.Jelena bought three kilos of apples. Add a pound of sugar. ''& `` The process of determining the amount of a substance in a given volume or area, typically using a measuring device such as a balance or a graduated container. '' \\

\hline
Commerce\_buy & `` These are words describing a basic commercial transaction involving a Buyer and a Seller exchanging Money and Goods, taking the perspective of the Buyer.  The words vary individually in the patterns of frame element realization they allow.  For example, the typical pattern for the verb BUY: Buyer buys Goods from Seller for Money. Abby bought a car from Robin for \$5,000. '' & `` The act of obtaining goods or services by giving something in exchange, such as money, in a transaction between a buyer and a seller, often in a retail setting. ''\\
\hline
\end{tabularx}
\caption{Sample FN 1.7 gold definitions vs. Llama-3.1-8B-Instruct generated definitions.}
\label{tab:frame-only-llama-generated}
\end{table*}

Table~\ref{tab:frame-only-llama-generated} presents a sample qualitative comparison between the FrameNet 1.7 gold defintions and the definitions generated by Llama-3.1-8B-Instruct. The results reveal both the strengths and limitations of the implicit semantic knowledge of the model. On the positive side, the model often succeeds in identifying the core intent of a frame: for example, Activity\_prepare is described as a process of preparation, and frame Commerce\_buy is summarized as a transactional exchange between a buyer and seller. These outputs suggest that the model has internalized broad semantic associations aligned with the FrameNet’s conceptual structure. However, the outputs also show clear shortcomings: they are often loosely structured, and sometimes hallucinated.~For example, Judicial\_body degenerates into a flat list of roles without institutional nuance, while Measure\_mass reduces to a generic account of measurement, missing the linguistic patterns emphasized in FrameNet. Taken together, these observations suggest that while Llama encodes latent semantic knowledge of frames, its outputs lack the precision, and role-structure sensitivity of curated FrameNet definitions. This gap highlights the challenge of moving from broad conceptual knowledge to the more fine-grained, lexically anchored semantics required for frame-semantic resources. Future work could explore controlled definition generation techniques that enforce conciseness and role-structure fidelity, hybrid approaches that combine gold and generated definitions to support under-resourced frames, and extend definition generation to multilingual settings where FrameNet-style resources remain scarce.

\begin{table*}[pht!]
    \centering
    \begin{tabularx}{\linewidth}{|p{0.30\linewidth}|X|}
        \hline
        \textbf{Sentence} & \textbf{Candidate Frames - Frame \& Lexical Unit definitions} \\
        \hline
         ``the government has also entered into \textcolor{blue}{new} cooperation agreements with several countries , most notably russia.'' & 
        \textcolor{red}{\textit{Familiarity}} - An Entity is presented as having been seen or experienced by a (typically generic and backgrounded) Cognizer on a certain number of...  - new.a: unfamiliar or strange to .
        \vspace{0.2em}\par
        \textcolor{darkgreen}{\textit{Age }} - An Entity has existed for a length of time, the Age. The Age can ... - new: not existing before \\
        \hline
        `` the u.s. economy may be on the verge of \textcolor{blue}{falling} back into recession after more than a year of half-hearted recovery that failed to generate either jobs or hope, according to economists.'' & 
        \textit{Conquering} - This frame describes a Theme losing its autonomy and perhaps sustaining material damage ... - fall: to be taken over and potentially destroyed by an army
        \vspace{0.2em}\par
        \textcolor{darkgreen}{\textit{Motion\_directional}} - In this frame a Theme moves in a certain Direction which is often determined by gravity ... - fall: move from a higher to a lower level, typically rapidly and without control
        \vspace{0.2em}\par
        \textcolor{red}{\textit{Change\_position\_on\_a\_scale}} -This frame consists of words that indicate the change of an Item's position on a scale... - fall: decrease. \\
        \hline
        `` qn: in which \textcolor{blue}{city} is john 's laptop on the evening of dec 11th ? '' & 
        \textcolor{darkgreen}{\textit{Locale\_by\_use}} - Geography as defined by its use.   'You must land in the next airfield, as th...- city: an inhabited place of greater size than a town or village
        
        \textcolor{red}{\textit{Political\_locales}} - This frame covers words that name Locations as defined politically... - city: a municipal centre incorporated by the state or province, or a town created a city by charter and containing a cathedral.\\
        \hline
        `` in late february 2003 , north korea restarted its 5ww(e) reactor , and in march , \textcolor{blue}{reports} indicated that technicians were active at the radiochemistry laboratory , and on 2 october , the north korean foreign ministry declared that the reprocessing of 8,000 spent fuel rods had been completed `` to increase its nuclear deterrent force . '' & 
        \textit{Reporting}- In this frame an Informer informs the Authorities of the illegal or otherwise improper Behavior of the Wrongdoer.... - report: \vspace{0.2em}\par
        \textcolor{darkgreen}{\textit{Text}} - A Text is an entity that contains linguistic, symbolic information on a Topic, created by an Author at the Time\_of\_creation ...
        - report: an account given of a matter after investigation or consideration.\vspace{0.2em}\par
        \textcolor{red}{\textit{Statement}} - This frame contains verbs and nouns that communicate the act of a Speaker to address a Message to some Addressee using language. Instead of (or in addition to) ... - report: an account given of a matter after investigation or consideration \\
        \hline
        ``estimates vary on how soon north korea could begin \textcolor{blue}{operating} a uranium enrichment plant , but north korea probably could not produce significant quantities of weapons-grade heu until the end of the decade.'' &  \textit{Military\_operation} - The Military of a Possessor (either a nation, institution, or private individual) conducts large-scale activities .... - operate: be militarily active or perform military actions.\vspace{0.2em}\par
        
        \textcolor{red}{\textit{Operating\_a\_system}} - An Operator manipulates the substructure of a System such that the System performs the function it was created for... - operate: control the functioning of (a device, system, or institution).\vspace{0.2em}\par
        
        \textit{Being\_in\_operation} - A Device or machine is in (or out of) service. Note that being broken or functional is... - operate: (of an artifact or machine) be active.\vspace{0.2em}\par
        
        \textcolor{darkgreen}{\textit{Using}} - An Agent manipulates an Instrument in order to achieve a Purpose. .. - operate: control a device (in order to acheive the prototypical function of the device).\\
        \hline
    \end{tabularx}
    \caption{Few more agreeing wrong predictions, where both FIDO and Llama-3.1-8B predict the same incorrect frame. Target words are marked in \textcolor{blue}{blue}, gold frames in \textcolor{darkgreen}{green}, and predicted frames in \textcolor{red}{red}.}
    \label{tab:agreeing-wrong-predictions}
\end{table*}

\onecolumn

\setlength{\arrayrulewidth}{0.5pt}  

\lstdefinestyle{prompt}{
  basicstyle=\ttfamily\scriptsize,
  breaklines=true,
  breakatwhitespace=false,
  columns=fullflexible,
  keepspaces=true,
  showstringspaces=false,
  frame=none,
  aboveskip=0pt,
  belowskip=0pt
}

\newcolumntype{L}[1]{>{\RaggedRight\arraybackslash}p{#1}}

\setlength{\LTpre}{0pt}
\setlength{\LTpost}{0pt}

\begin{table}[t]
\begin{small}
\renewcommand{\arraystretch}{1.05}
\centering
\begin{tabular}{| L{0.20\linewidth} | L{0.70\linewidth} |}
\hline
\textbf{Task} & \textbf{Prompt} \\
\hline

\raggedright Simple Prompt &
\begin{lstlisting}[style=prompt]
You are an expert in FrameNet semantics.

Your task is to identify the most appropriate FrameNet frame that best captures the meaning of a given target word in context.

You will be given:
- A sentence containing the target word.
- Target Word
- A list of frames along with their descriptions.

Your output must be a **single JSON object** in this exact format:
{"frame_Name": "Intentionally_act"}

Where:
- "frame_Name" is the exact name of the selected FrameNet frame.

Sentence: additionally , over the years , syria has solicited proposals from other countries including argentina , india,
and italy.
Target Word: country

Frames:
Locale_by_use ; Lexical Unit Definition : country.n: districts outside large urban areas.
Political_locales ; Lexical Unit Definition : country.n: a nation with its own government, occupying a particular territory.

Which of the given frames best represents the meaning of the target word country in the sentence above?
\end{lstlisting}
\\ \hline

\raggedright Direct QA Prompt &
\begin{lstlisting}[style=prompt]
You are an expert in FrameNet semantics.

Your task is to identify the most appropriate FrameNet frame that best captures the meaning of a given target word in context.

You will be given:
- A sentence containing the target word.
- Target Word
- A list of frame options labeled A, B, C, etc., along with their descriptions.

Your output must be a **single JSON object** in this exact format:
{"frame_Option": "C", "frame_Name": "Intentionally_act"}

Where:
- "frame_Option" is the correct option letter.
- "frame_Name" is the exact name of the selected FrameNet frame.

Do NOT include any explanation, comments, or extra text.
Only return the JSON object.

Sentence: additionally , over the years , syria has solicited proposals from other countries including argentina , india,
and italy.
Target Word: country
The different senses of this word are
1. country.n: districts outside large urban areas
2. country.n: a nation with its own government, occupying a particular territory.
These senses can be related to the frames: 'Locale_by_use', 'Political_locales' respectively
Which of the following frames best represents the meaning of the target word country in the sentence above?
Options:
A. Frame: Locale_by_use
B. Frame: Political_locales
\end{lstlisting}
\\ \hline

\end{tabular}
\caption{Prompts used in the Simple and Direct QA experiments.}
\label{tab:prompt-simple-directqa}
\end{small}
\end{table}

\begin{table}[t]
\renewcommand{\arraystretch}{1.05}
\centering
\begin{tabular}{| L{0.20\linewidth} | L{0.70\linewidth} |}
\hline
\textbf{Task} & \textbf{Prompt} \\
\hline
\raggedright Artifacts Prompt &
\begin{lstlisting}[style=prompt]
You are an expert in FrameNet and artifact semantics.

Your task is to select the most appropriate FrameNet frame that best represents the prototypical function of a given artifact.

Definitions:
- The Prototypical Function refers to the core activity or process that the artifact is typically used to perform.
- An Artifact refers to a human-made object that has a specific purpose or function.
- Choose "None of above" (Option 43) if none of the frames meaningfully represent the core function of the artifact.

Artifact: abacus
Definition: a tablet placed horizontally on top of the capital of a column as an aid in supporting the architrave.

Frame Options:
1. Frame: Cause_motion
2. Frame: Cause_to_be_dry
3. Frame: Excreting
4. Frame: Containing
5. Frame: Cause_harm
6. Frame: Rite
7. Frame: Protecting
8. Frame: Building
9. Frame: Education_teaching
10. Frame: Cutting
11. Frame: Cooking_creation
12. Frame: Light_movement
13. Frame: Bringing
14. Frame: Dimension
15. Frame: Closure
16. Frame: Hunting
17. Frame: Supporting
18. Frame: Agriculture
19. Frame: Cure
20. Frame: Competition
21. Frame: Commercial_transaction
22. Frame: Cause_to_fragment
23. Frame: Cause_fluidic_motion
24. Frame: Eclipse
25. Frame: Grooming
26. Frame: Make_noise
27. Frame: Cause_temperature_change
28. Frame: Ingestion
29. Frame: Create_representation
30. Frame: Inhibit_movement
31. Frame: Residence
32. Frame: Performing_arts
33. Frame: Setting_fire
34. Frame: Attaching
35. Frame: Removing
36. Frame: Wearing
37. Frame: Sleep
38. Frame: Contacting
39. Frame: Self_motion
40. Frame: Perception_experience
41. Frame: Text_creation
42. Frame: Reading_activity
43. Frame: None of above

Pick the best option (1/2/3/.../43):

Answer:
\end{lstlisting}
\\ \hline
\end{tabular}
\caption{Prompt used in the Artifacts experiment.}
\label{tab:prompt-artifacts}
\end{table}

\begin{table}[t]
\begin{small}
\renewcommand{\arraystretch}{1.05}
\centering
\begin{tabular}{| L{0.20\linewidth} | L{0.70\linewidth} |}
\hline
\textbf{Task} & \textbf{Prompt} \\
\hline
\raggedright QA Fine-tuning \& YAGS Prompt &
\begin{lstlisting}[style=prompt]
Select the most appropriate frame that matches the meaning of the target word in the sentence. (This is a frame semantic parsing task.)
Target word: "complex"
Sentence: North Korea established a nuclear energy research complex at Yongbyon in 1964 and set up a Soviet research reactor at the site in mid-2002.
Options:
A. Frame: Locale_by_use - Geography as defined by use ; Lexical Unit Definition : complex.n - a group of similar buildings or facilities on the same site.
B. Frame: System - A Complex formed out of Component_entities with a particular Function ; Lexical Unit Definition : complex.n - an interlinked system; a network.
Pick the best option (A/B).
Answer:
\end{lstlisting}
\\ \hline
\end{tabular}
\caption{Prompt used in the QA fine-tuning and YAGS experiment.}
\label{tab:prompt-qa-finetune}
\end{small}
\end{table}

\begin{table}[t]
\renewcommand{\arraystretch}{1.05}
\centering
\begin{tabular}{| L{0.20\linewidth} | L{0.70\linewidth} |}
\hline
\textbf{Task} & \textbf{Prompt} \\
\hline
\raggedright Frame Definition Generation &
\begin{lstlisting}[style=prompt]
You are a semantic and FrameNet expert who defines FrameNet frames. Provide a clear and concise definition of the given frame.

Guidelines: 
- Provide a definition that explains the situation or event the frame describes (not just a dictionary meaning).
- Describe the typical scenario or context that the frame captures.
- Include the main participants or roles (Frame Elements) only if necessary to clarify the definition.
- Do NOT give usage examples, paraphrases, or lexical units.
- Keep the definition self-contained and precise.

Respond only with the definition in JSON format as shown below

Format:
{{ "frame": "<frame_name>", "definition": "<definition>" }}

What is the definition of the FrameNet frame "{frame_name}" ?
\end{lstlisting}
\\ \hline

\raggedright Evaluation of Generated Frame Definitions &
\begin{lstlisting}[style=prompt]
You are an expert in FrameNet semantics.

Your task is to identify the most appropriate FrameNet frame definition that best captures the meaning of a given target word in context.

You will be given:
- A sentence containing the target word.
- Target Word
- A list of frame definition options without frame names explicitly, labeled as A, B, C, etc.

Your output must be a **single JSON object** in this exact format:
{{"frame_definition_Option": "C"}}

Where:
- "frame_definition_Option" is the correct option letter of the frame definition.

Do NOT include any explanation, comments, or extra text.
Only return the JSON object.

Sentence: your contribution to goodwill will mean more than you may know.

Target Word: know

Which of the following frame definitions best represents the meaning of the target word know in the sentence above? 
Frame Definitions :
A. A situation where one person or entity is well-acquainted or knowledgeable about another, often as a result of past interactions or shared experiences.
B. A state of being confident or certain about the truth or existence of something, often resulting from evidence or reasoning.
C. A state of being informed or knowledgeable about a particular fact, situation, or issue, often involving a recognition of a problem or a potential threat, and a sense of responsibility or obligation to take action or address the situation.
D. A process or activity that highlights the distinctions or differences between two or more entities, concepts, or categories, often for the purpose of establishing distinct identities or unique characteristics.
\end{lstlisting}
\\ \hline
\end{tabular}
\caption{Prompts used for frame definition generation and evaluation of generated frame definitions.}
\label{tab:prompt-def-gen-probing}
\end{table}

\end{document}